\documentclass[numbers]{article}

\usepackage{color,soul}

\usepackage[final]{neurips_2024}

\usepackage[utf8]{inputenc} 
\usepackage[T1]{fontenc}    
\usepackage{hyperref}       
\usepackage{url}            
\usepackage{booktabs}       
\usepackage{amsfonts}       
\usepackage{nicefrac}       
\usepackage{microtype}      
\usepackage{xcolor}         
\usepackage{wrapfig}
\usepackage{graphicx}
\usepackage{natbib}
\usepackage{amsmath}
\usepackage{amsthm}
\usepackage{array}    
\usepackage[english]{babel}

\newtheorem{lemma}{Lemma}
\newtheorem{definition}{Definition}
\NewDocumentCommand{\shuiwang}
{ mO{} }{\textcolor{blue}{\textsuperscript{\textit{Shuiwang Ji}}\textsf{\textbf{\small[#1]}}}}
\NewDocumentCommand{\heng}
{ mO{} }{\textcolor{red}{\textsuperscript{\textit{Heng}}\textsf{\textbf{\small[#1]}}}}
\NewDocumentCommand{\xiaofeng}
{ mO{} }{\textcolor{cyan}{\textsuperscript{\textit{Xiaofeng}}\textsf{\textbf{\small[#1]}}}}

\title{Invariant Tokenization of Crystalline Materials for Language Model Enabled Generation}

\author{%
  Keqiang Yan$^{1}$, Xiner Li$^{1}$, Hongyi Ling$^{1}$, Kenna Ashen$^{1}$,  Carl Edwards$^{2}$, \\
  \textbf{Raymundo Arróyave}$^{1}$, \textbf{Marinka Zitnik}$^{3}$, \textbf{Heng Ji}$^{2}$, \textbf{Xiaofeng Qian}$^{1}$, \textbf{Xiaoning Qian}$^{1}$, \\
  \textbf{Shuiwang Ji}$^{1}$ \\
  $^{1}$Texas A\&M University, College Station, TX 77843, USA \\
  $^{2}$University of Illinois Urbana-Champaign, Champaign, IL 61820, USA \\
  $^{3}$Harvard University, Boston, MA 02115, USA \\
  \texttt{\{keqiangyan,lxe,hongyiling,kashen13,rarroyave,feng,xqian,sji\}@tamu.edu} \\
   \texttt{\{cne2,hengji\}@illinois.edu} \\
   \texttt{marinka@hms.harvard.edu} 
}

\begin{document}

\maketitle

\begin{abstract}

We consider the problem of crystal materials generation using language models~(LMs). A key step is to convert 3D crystal structures into 1D sequences to be processed by LMs. Prior studies used the crystallographic information framework~(CIF) file stream, which fails to ensure $SE(3)$
and periodic invariance and may not lead to unique sequence representations for a given crystal structure. Here, we propose a novel method, known as Mat2Seq, to tackle this challenge. Mat2Seq converts 3D crystal structures into 1D sequences and ensures that different mathematical descriptions of the same crystal are represented in a single unique sequence, thereby provably achieving $SE(3)$ and periodic invariance. Experimental results show that, with language models, Mat2Seq achieves promising performance in crystal structure generation as compared with prior methods.
\end{abstract}

\section{Introduction}

Discovering crystalline materials with desired properties is valuable for advancing a variety of technological sectors~\citep{petousis2016benchmarking, mahapatra2021piezoelectric,wang2019ferroelectric,xiao2020berrymemory,wang2020nlpc, choudhary2024jarvis,zhang2023artificial,yan2022periodic,potnet}. However, current materials discovery relies on trial-and-error wet-lab experimental methods that are time-consuming and expensive. Computational methods based on advanced quantum mechanical approaches such as first-principles density functional theory~(DFT) have sped up this process in the last two decades. Nevertheless, the high computational cost of these methods ranging from $O(n^3)$ to $O(n^7)$, where $n$ is number of electrons in a material system, poses a significant challenge for high-throughput screening of the infinite materials space. 

Recent advances in large language models (LLMs) have showcased substantial capabilities in real-time question answering and image generation. However, the application of these models to atomistic systems, including molecular and material structures, remains relatively underexplored. Specifically, while recent studies have attempted to represent molecules and crystals using XYZ and CIF structure file formats~\cite{flam2023language, gruver2024fine, antunes2023crystal}, they have not adequately addressed the critical issues of uniqueness and invariance in sequence representations of 3D crystal structures, which play key roles for successful LLM-based discovery of novel crystalline materials with discussions provided in Sec.~\ref{sec:related_llm}.

The concept of uniqueness~\cite{wang2022comenet, comformer} in crystal sequences mandates a bijective relationship between crystal structures and their sequence representations. Current approaches using CIF files directly in crystal LLMs violate this uniqueness criterion, leading to significantly varied sequences for identical crystals, particularly in terms of atom ordering and fractional coordinates within the unit cell as illustrated in Figure~\ref{fig:limits_cif}. This lack of unique sequence representations means that an infinite number of CIF files can describe the same crystal, necessitating extensive augmentation during LLM training to recognize these equivalences. Without such augmentation, previous CIF file based methods \cite{flam2023language, gruver2024fine} essentially train crystal LLMs using only a limited set of crystal structure snapshots.

In this work, we develop a framework for creating unique and complete crystal sequence representations, followed by the construction of a material LLM capable of generating novel crystal structures with desired properties of interest.
To accomplish this, several challenges must be addressed. First, unlike molecular structures that contain a finite number of atoms, each crystal structure consists of an infinite number of atoms through a periodic translation of a finite set of atoms in a unit cell 
along the direction of three lattice vectors in three-dimensional (3D) space. Consequently, there are numerous different unit cells for the same crystal as shown in Figure~\ref{fig:limits_cif}.
A unique and invariant unit cell must therefore be selected for each crystal. Second, it is crucial that this unit cell can be represented in a one-dimensional (1D) sequence that maintains invariance under arbitrary rotations and ensures completeness, allowing the full reconstruction of the crystal structure from its sequence representation. To tackle these complexities, we introduce Mat2Seq, a method that systematically transforms 3D crystal structures into 1D sequences. This is achieved by first identifying $SO(3)$ equivariant unit cells and subsequently converting these into $SE(3)$ invariant sequences. By integrating these unique and complete sequences into LLMs, we develop conditional generation capabilities for novel crystal structures. Our experimental results in crystal structure prediction and crystal discovery with desired properties validate the efficacy of Mat2Seq.

\section{Preliminaries and related work}
\label{sec:related_work}

\subsection{Crystal structure generation}

In this work, we study the problem of generating 3D crystals from scratch. Unlike small molecules, crystalline materials are defined by a unit cell which contains a set of atoms repeated infinitely across three-dimensional space along three periodic lattice vectors. Following notations of ComFormer~\cite{comformer}, each crystal structure is represented by $\mathbf{M} = (\mathbf{A}, \mathbf{P}, \mathbf{L})$, where $\mathbf{A}=[\boldsymbol{a}_1, \boldsymbol{a}_2, \cdots, \boldsymbol{a}_n] \in \mathbb{R}^{d_a \times n}$ represents the $d_a$-dimensional feature vectors of $n$ atoms within the unit cell, $\textbf{P} = [\boldsymbol{p}_1, \boldsymbol{p}_2, \cdots, \boldsymbol{p}_n] \in \mathbb{R}^{3 \times n}$ represents the 3D Euclidean positions of these $n$ atoms, and $\textbf{L} = [\boldsymbol{\ell}_1, \boldsymbol{\ell}_2, \boldsymbol{\ell}_3] \in \mathbb{R}^{3\times 3}$ specifies three periodic lattice vectors, representing the repeating patterns of the unit cell in 3D space. The infinite structure of a given crystal $\mathbf{M} = (\mathbf{A}, \mathbf{P}, \mathbf{L})$ is formalized as
$    \hat{\mathbf{P}} =  \{\hat{\boldsymbol{p}_i} | \hat{\boldsymbol{p}_i} = \boldsymbol{p}_i + k_1\boldsymbol{\ell}_1 + k_2\boldsymbol{\ell}_2 + k_3\boldsymbol{\ell}_3,~ k_1, k_2, k_3 \in \mathbb{Z}, 
    i \in \mathbb{Z}, 1 \le i \le n \}, 
    \hat{\mathbf{A}} =  \{\hat{\boldsymbol{a}_i} | \hat{\boldsymbol{a}_i} = \boldsymbol{a}_i, i \in \mathbb{Z}, 1 \le i \le n \},$
where $\hat{\mathbf{P}}$ denotes the positions of atoms and their infinite repeats in the 3D space, and $\hat{\mathbf{A}}$ denotes the corresponding feature vectors. This work targets two primary tasks:

\textbf{Crystal structure generation}: Using a dataset of 3D crystals $\{\mathbf{M}_j\}_{j=1}^m$, we aim to develop a structure generative model $p_\theta (\cdot|\mathbf{A}_j)$ to synthesize valid and stable 3D crystal structures for input compositions.

\textbf{Conditional generation}: With a dataset $\{(\mathbf{M}_j, s_j)\}_{j=1}^m$ , where $s_j$ denotes a specific property of $\mathbf{M}_j$, we aim to establish a conditional generative model $p_\theta (\cdot|s)$ to generate  3D crystal structures possessing the property $s$.

Generative models for crystal structures must address complex geometric requirements, such as ensuring periodic and unit cell $SE(3)$ invariance~\cite{yan2022periodic,cdvae,comformer,luo2024towards}. The ideal models should consider $\mathbf{M} = (\mathbf{A}, \mathbf{P}, \mathbf{L})$ and its equivalents under periodic transformations or rotations and translations as identically probable~\cite{cdvae,comformer}.
CDVAE~\cite{cdvae} achieves periodic and $SE(3)$ invariance by encoding crystal structures into $SE(3)$ invariant latent features but faces limitations in restricting the loss function to be periodic invariant. This limitation is later rectified in SyMat~\cite{luo2024towards}. Beyond this, DiffCSP~\cite{jiao2024crystal} is specifically optimized for crystal structure prediction that generates stable crystal structures for given compositions. In contrast, MatterGen~\cite{zeni2023mattergen} models $(\mathbf{A}, \mathbf{P}, \mathbf{L})$ in an equivariant manner by gradually corrupting them into known distributions.

While the generation of 3D crystal structures as 3D point clouds is well-established, applying powerful language models to this task is novel and nontrivial as discussed in Sec.~\ref{sec:related_llm}. Our approach uniquely transforms infinite 3D crystal structures into $SE(3)$ and periodic invariant sequences, demonstrating that language models can effectively generate crystal structures with high performance.

\subsection{Language model for 3D crystal structures}
\label{sec:related_llm}

\begin{figure}[t]
    \centering\includegraphics[width=0.99\linewidth]{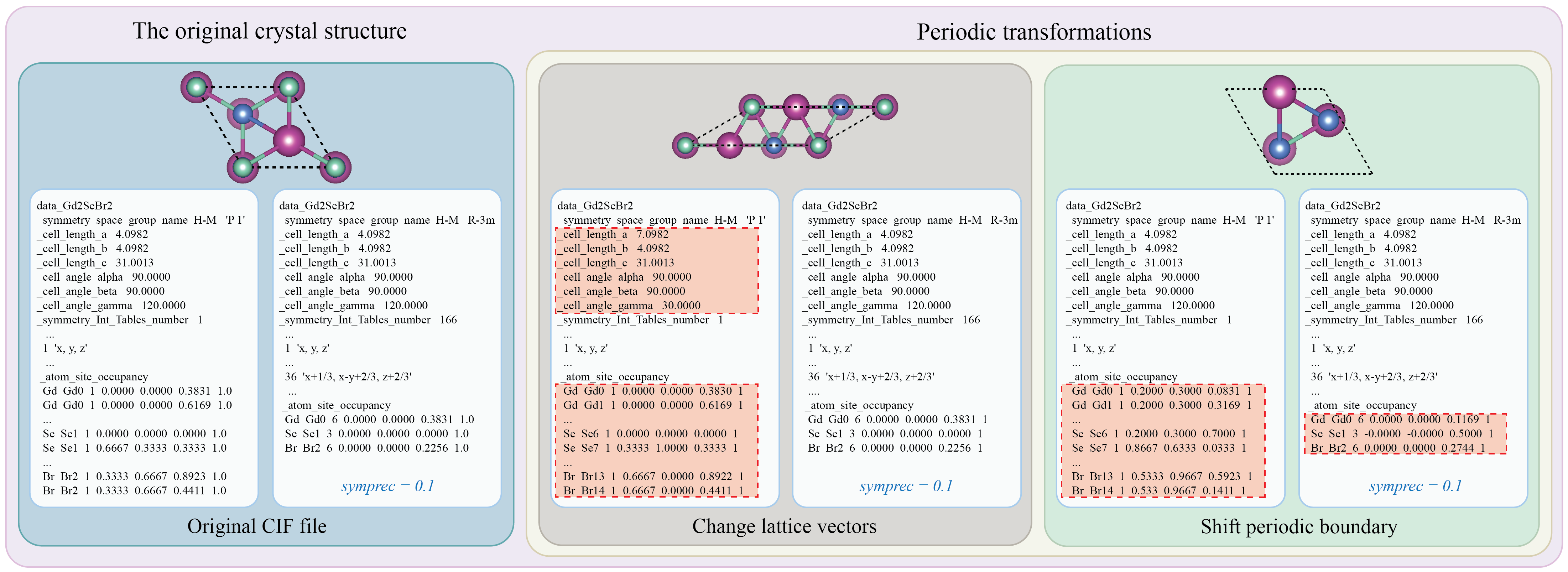}
    \caption{Limitations of directly using CIF files in achieving unique crystal sequence representations. This figure demonstrates variations in CIF files, either with or without symmetry control command denoted as "symprec", for the same crystal structure subjected to periodic transformations. Changes in the CIF contents are highlighted in red. Periodic transformations can significantly alter the unit cell structures, resulting in distinct CIF files \emph{differed by fractional coordinates, atom ordering, and lattice parameters} for the same underlying crystal.
    }
    \vspace{-3mm}
    \label{fig:limits_cif}
\end{figure}

Enabling language models to generate 3D crystal structures necessitates transforming these structures into sequence representations. Several studies have employed CIF files as sequence inputs for this purpose~\cite{flam2023language, gruver2024fine, antunes2023crystal}. These approaches convert a set of 3D crystals $\{\mathbf{M}_j\}_{j=1}^m$ into a dataset of sequences, $\boldsymbol{C} = \{C_1, C_2, \cdots, C_m\}$, where each $C_j$ represents the CIF text of $\mathbf{M}_j$. The sequence $C_j=\{c_1, c_2, \cdots, c_{n_j}\}$ of length $n_j$ are composed of tokens $c_i$ from a predefined vocabulary $V$. Autoregressive language models are either trained~\cite{flam2023language, antunes2023crystal} or fine-tuned~\cite{gruver2024fine} to encode these sequences by maximizing the conditional probabilities of each token given its predecessors, $p(C_i|\theta)=\prod_{j=1}^{n_i}p(c_j | c_1:c_{j-1}; \theta)$, with the objective to maximize the probabilities of the dataset $p(\boldsymbol{C}|\theta)=\prod_{i=1}^{m}p(C_i; \theta)$. During generation, novel crystal sequences are produced from $p(\theta)$ in an autoregressive manner. 

However, ideal crystal generative models should consider $\mathbf{M} = (\mathbf{A}, \mathbf{P}, \mathbf{L})$ and its equivalents under periodic transformations or rotations and translations as identically probable, a criterion unmet by current models due to the variability in CIF text descriptions for the same crystal. For example, there are different sequence representations $C_i=\{c_1, c_2, \cdots, c_{n_i}\}$ and $C_i'=\{c_1', c_2', \cdots, c_{n_i'}'\}$ for the same crystal, yet ideally, $\prod_{j=1}^{n_i}p(c_j | c_1:c_{j-1}; \theta) = \prod_{j=1}^{n_i'}p(c_j' | c_1':c_{j-1}'; \theta)$ should hold. Despite being highly non-trivial, this is overlooked by previous studies. 
This issue is illustrated through the failure cases of CIF-based crystal language models~\cite{flam2023language, gruver2024fine, antunes2023crystal} in Figure~\ref{fig:limits_cif}.

\textbf{Differences with previous works}. It is crucial to recognize that prior works~\cite{flam2023language, gruver2024fine, antunes2023crystal} using CIF files as input violate invariance and generate a multitude of different sequence representations for the same crystal. Consequently, these models fail to consistently recognize $\mathbf{M} = (\mathbf{A}, \mathbf{P}, \mathbf{L})$ and its equivalents under periodic transformations including shifting periodic boundaries and changing lattice vectors as identically probable. While intensive data augmentation can mitigate this issue, it significantly burdens the training process of language models. In contrast, our Mat2Seq approach maps all equivalent crystals to a unique sequence, naturally ensuring that all equivalents are considered equally probable without the need for data augmentation.

\section{Tokenization of 3D crystal materials}
\label{sec:method}

\begin{figure}[t]
    \centering\includegraphics[width=0.99\linewidth]{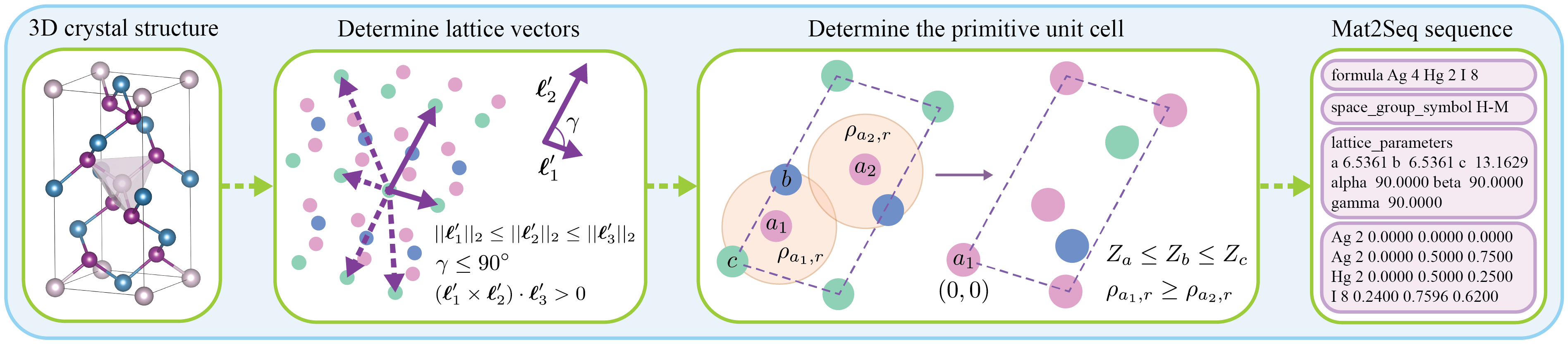}\vspace{-0.3cm}
    \caption{The pipeline of Mat2Seq that converts 3D crystal structures into unique crystal sequences. Mat2Seq first determines $SO(3)$ equivariant and periodic invariant lattice vectors using Niggli cell reduction~\cite{shi2022niggli}, then determines the primitive unit cell. After that, Mat2Seq converts the determined $SO(3)$ equivariant and periodic invariant primitive cells into $SE(3)$ and periodic invariant sequences.
    }
    \vspace{-3mm}
    \label{fig:mat2seq_pip}
\end{figure}

In this section, we demonstrate \textbf{Mat2Seq} for transforming 3D crystal structures into 1D sequences that adhere to the principles of $SE(3)$ invariance, periodic invariance, and completeness. We first discuss requirements for ideal crystal sequence representations in Sec.~\ref{sec:req} and provide formal definitions of unit cell $SE(3)$ invariance, periodic invariance, and completeness. We then show how to design crystal sequence representations to achieve these requirements in Sec.~\ref{sec:SO3_cell} and~\ref{sec:SE3_seq}, with the pipeline shown in Figure~\ref{fig:mat2seq_pip}. We then provide property proofs of \textbf{Mat2Seq} in Sec.~\ref{sec:proofs}, and demonstrate the combination of \textbf{Mat2Seq} and language models for crystal structure generation in Sec.~\ref{sec:combine_LMs}.

\subsection{Requirements for ideal crystal sequence representations}
\label{sec:req}

\textbf{Uniqueness}. Ideal crystal sequences need to satisfy uniqueness which ensures a deterministic mapping of a crystal structure $\mathbf{M} = (\mathbf{A}, \mathbf{P}, \mathbf{L})$ and its equivalents, as transformed by periodic and $SE(3)$ operations, to a singular, unique sequence. However, to the best of our knowledge, none of previous studies~\cite{flam2023language, gruver2024fine, antunes2023crystal} achieves uniqueness for crystal sequence representations.

When a sequence inherently satisfies this criterion, it guarantees that all structural equivalents are recognized as equally probable by language models, thus eliminating the need for any manual adjustments for language models or extensive augmentation during the training phase. Following ComFormer~\cite{comformer}, we further define the necessary invariances for ideal sequence representations.

\begin{definition}[Unit Cell $SE(3)$ Invariance]
A function $f:(\mathbf{A},\mathbf{P}, \mathbf{L}) \to \mathcal{X}$ is unit cell $SE(3)$ invariant if, for any rotation transformation $\bold{R} \in \mathbb{R}^{3 \times 3}, |\bold{R}|= 1$ and translation transformation $\bold{b} \in \mathbb{R}^{3}$ , we have $f(\mathbf{A}, \mathbf{P}, \mathbf{L}) = f(\mathbf{A}, \bold{R}\mathbf{P}+b, \bold{R}\mathbf{L})$. 
\end{definition}

\begin{definition}[Periodic Invariance]
A function $f:(\mathbf{A},\mathbf{P}, \mathbf{L}) \to \mathcal{X}$ is periodic invariant if, for any possible minimum unit cell representations $\mathbf{M}' = (\mathbf{A}', \mathbf{P}', \mathbf{L}')$ representing a given infinite crystal structure $(\hat{\mathbf{P}}, \hat{\mathbf{A}})$, we have $f(\mathbf{A}, \mathbf{P}, \mathbf{L}) = f(\mathbf{A}', \mathbf{P}', \mathbf{L}')$.
\end{definition}

This definition captures two periodic transformations that result in distinct minimum unit cell representations of the same crystal structure: (1) shifting periodic boundaries, and (2) altering periodic patterns while maintaining the same unit cell volume as shown in Figure~\ref{fig:limits_cif}.

\textbf{Completeness}. Ideal crystal sequences need to be complete, allowing the full reconstruction of 3D crystal structures from their 1D sequences.

\subsection{Determination of $SO(3)$ equivariant unit cells}
\label{sec:SO3_cell}

We propose Mat2Seq that converts 3D crystal materials into 1D sequences by (1) determining the $SO(3)$ equivariant unit cells and (2) converting $SO(3)$ equivariant unit cells into $SE(3)$ invariant sequences. In this section, we demonstrate how to determine the $SO(3)$ equivariant unit cell of a given crystal structure $\mathbf{M} = (\mathbf{A}, \mathbf{P}, \mathbf{L})$.

\textbf{Determination of lattice vectors}. Let's start with an arbitrary minimum (primitive) cell with vectors $\boldsymbol{\ell}_1, \boldsymbol{\ell}_2, \boldsymbol{\ell}_3$. A set of $SO(3)$ equivariant lattice vectors can be determined by Niggli cell reduction~\cite{shi2022niggli} as follows. The three lattice vectors $\boldsymbol{\ell}_1', \boldsymbol{\ell}_2', \boldsymbol{\ell}_3'$ of a new primitive cell can be represented in terms of the original ones as $\boldsymbol{\ell}_i' = k_{i,1}\boldsymbol{\ell}_1 + k_{i,2}\boldsymbol{\ell}_2 + k_{i,3}\boldsymbol{\ell}_3,~ k_{i,1}, k_{i,2}, k_{i,3} \in \mathbb{Z}$. Then, a unique and $SO(3)$ equivariant cell can be determined by choosing three shortest non-planar vectors $\boldsymbol{\ell}_1', \boldsymbol{\ell}_2', \boldsymbol{\ell}_3'$ that form a right-hand system using three Euclidean vector lengths and three relative angles (\textit{i.e.} six lattice parameters)~\cite{comformer,shi2022niggli}.

\textbf{Determination of primitive unit cells}. After lattice vectors $\mathbf{L}_u = [\boldsymbol{\ell}_1', \boldsymbol{\ell}_2', \boldsymbol{\ell}_3']$ have been determined, a unique unit cell is determined as follows. For different unit cells with the same lattice vectors $\boldsymbol{\ell}_1', \boldsymbol{\ell}_2', \boldsymbol{\ell}_3'$, they differ by the origin of the cell. We determine the origin of unit cell by: \textbf{1.} smallest atomic number; \textbf{2.} local densities of atoms in the cell that are E(3) invariant measurements defined by 
    $\rho_{i, r} = \sum_{j \in \mathcal{N}(i, r)} Z_j$,
where $\mathcal{N}(i, r)$ denotes the neighbors of atom $i$ within the radius $r$, and $Z_j$ denotes the atomic number of neighboring atom $j$;
and \textbf{3.} densities along three lattice vectors $\boldsymbol{\ell}_1', \boldsymbol{\ell}_2', \boldsymbol{\ell}_3'$ that are $SE(3)$ invariant measurements defined as the following,
\begin{equation}
    \rho^{\boldsymbol{\ell}_1'}_{i, r} = \sum_{j \in \mathcal{N}^{\boldsymbol{\ell}_1'}(i, r)} Z_j, \rho^{\boldsymbol{\ell}_2'}_{i, r} = \sum_{j \in \mathcal{N}^{\boldsymbol{\ell}_2'}(i, r)} Z_j,\rho^{\boldsymbol{\ell}_3'}_{i, r} = \sum_{j \in \mathcal{N}^{\boldsymbol{\ell}_3'}(i, r)} Z_j,
\end{equation}
where $\mathcal{N}^{\boldsymbol{\ell}_1'}(i, r), \mathcal{N}^{\boldsymbol{\ell}_2'}(i, r), \mathcal{N}^{\boldsymbol{\ell}_3'}(i, r)$ denote the neighbors of atom $i$ within the radius $r$ along $\boldsymbol{\ell}_1', \boldsymbol{\ell}_2', \boldsymbol{\ell}_3'$ that satisfy $\boldsymbol{p}_{\text{frac}, j}^x > \boldsymbol{p}_{\text{frac}, i}^x, \boldsymbol{p}_{\text{frac}, j}^y > \boldsymbol{p}_{\text{frac}, i}^y,$ and $ \boldsymbol{p}_{\text{frac}, j}^z > \boldsymbol{p}_{\text{frac}, i}^z$, correspondingly. Here, $\boldsymbol{p}_{\text{frac}, j}$ denotes the fractional coordinate of atom $j$ as follows,
\begin{equation}
    \boldsymbol{p}_{j} = \boldsymbol{p}_{\text{frac}, j}^x \boldsymbol{\ell}_1' + \boldsymbol{p}_{\text{frac}, j}^y \boldsymbol{\ell}_2' + \boldsymbol{p}_{\text{frac}, j}^z \boldsymbol{\ell}_3'.
\end{equation}
After the unique atom $i$ is chosen by \textbf{1.}, \textbf{2.}, and \textbf{3.}, the primitive unit cell is determined by moving this atom to the origin through the following transformation applied to all other fractional coordinates,
\begin{align}
    {\boldsymbol{p}_{\text{frac}, j}^x}' = (\boldsymbol{p}_{\text{frac}, j}^x - \boldsymbol{p}_{\text{frac}, i}^x)~ \text{mod}~1, \\
    {\boldsymbol{p}_{\text{frac}, j}^y}' = (\boldsymbol{p}_{\text{frac}, j}^y - \boldsymbol{p}_{\text{frac}, i}^y)~ \text{mod}~1, \\
    {\boldsymbol{p}_{\text{frac}, j}^z}' = (\boldsymbol{p}_{\text{frac}, j}^z - \boldsymbol{p}_{\text{frac}, i}^z)~\text{mod}~1,
\end{align}
where the resultant $\mathbf{P}_{\text{frac}, u} = [\boldsymbol{p}_{\text{frac}, 1}', \boldsymbol{p}_{\text{frac}, 2}', \cdots, \boldsymbol{p}_{\text{frac}, n}'] \in \mathbb{R}^{n \times 3}$, and $\mathbf{P}_{u} =  \mathbf{L}_u \cdot \mathbf{P}_{\text{frac}, u}$.

\textbf{Remarks}. By determining lattice vectors and primitive unit cells, a $SO(3)$ equivariant and periodic invariant unit cell $(\mathbf{A}_u, \mathbf{P}_u, \mathbf{L}_u)$ is determined for a given crystal material, which means after applying arbitrary $SO(3)$ transformations $\bold{R} \in \mathbb{R}^{3 \times 3}, |\bold{R}|= 1$ with $\bold{b} \in \mathbb{R}^{3}$ and periodic transformations, the resultant unit cell will transform to $(\mathbf{A}_u, \mathbf{R}\mathbf{P}_u, \mathbf{R}\mathbf{L}_u)$. We then demonstrate how to construct the $SE(3)$ invariant crystal sequence representation based on $(\mathbf{A}_u, \mathbf{P}_u, \mathbf{L}_u)$ in the following section.

\subsection{$SE(3)$ invariant crystal sequence representations}
\label{sec:SE3_seq}

\begin{wrapfigure}{r}{0.5\textwidth}
\vspace{-0.4cm}
    \centering
\includegraphics[width=0.45\textwidth]{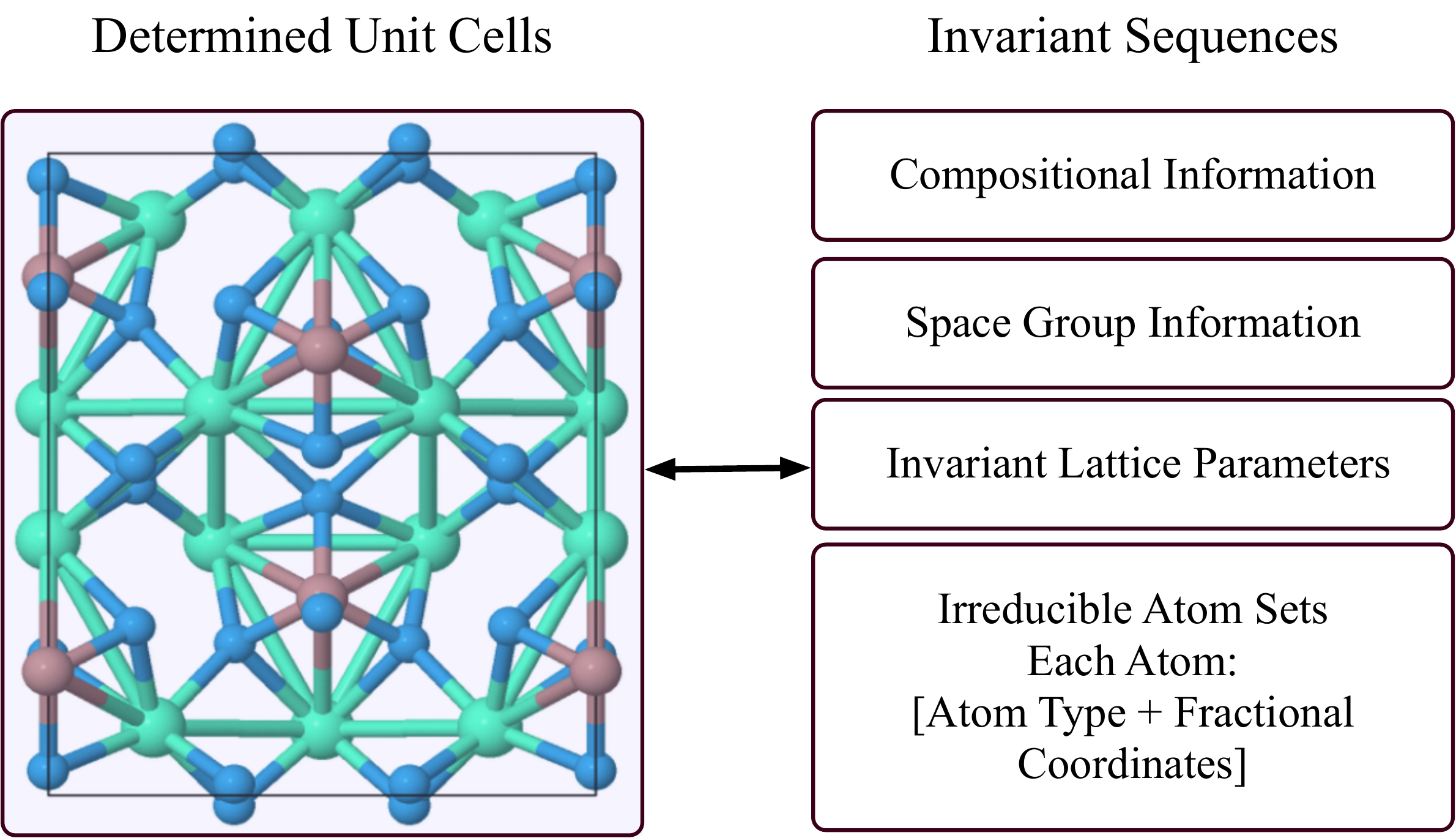}\vspace{-0.2cm}
    \caption{Converting determined unit cells into invariant crystal sequences.}
    \label{fig:mat-2}
\vspace{-3mm}
\end{wrapfigure}

Given determined $SO(3)$ equivariant and periodic invariant unit cells $\mathbf{M}=(\mathbf{A}_u, \mathbf{P}_u, \mathbf{L}_u)$, we aim to represent them by $SE(3)$ and periodic invariant sequences that are complete to guarantee the full reconstruction of crystal structures. As discussed in Sec.~\ref{sec:related_llm}, a fundamental requirement for crystal sequence representations is uniqueness which stipulates that (1) different crystal structures must correspond to distinct sequences and (2) a given crystal should yield the same sequence representation across all possible structural descriptions $\mathbf{M} = (\mathbf{A}, \mathbf{P}, \mathbf{L})$. Previous works~\cite{flam2023language, gruver2024fine, antunes2023crystal} directly use CIF files that breaks the uniqueness as shown in Figure~\ref{fig:limits_cif} as the input sequence for LMs. Therefore, these methods require tremendous augmentation effort during the training process to make LMs aware that these different sequences refer to the same crystal structure. Additionally, due to the $O(n^2)$ computational cost of decoder-only language models where $n$ is the window length, the sequence representations of crystal structures need to be efficient to support shorter window length and ease the burden of training and inference processes.

Rather than relying on CIF files, our approach extracts complete geometric information from the uniquely determined $SO(3)$ equivariant unit cells $\mathbf{M}=(\mathbf{A}_u, \mathbf{P}_u, \mathbf{L}_u)$. We first follow CIF files and use six invariant lattice parameters including $a =||\boldsymbol{\ell}_1||_2 ,b=||\boldsymbol{\ell}_2||_2,c=||\boldsymbol{\ell}_3||_2$ and $\alpha, \beta, \gamma$ that are three bond angles between $\boldsymbol{\ell}_1, \boldsymbol{\ell}_2, \boldsymbol{\ell}_3$ to represent $\mathbf{L}_u$ as invariant sequences.  Subsequently, we incorporate space group information of the crystal structures to reduce sequence length, documenting the \textbf{irreducible atom sets} alongside the corresponding symmetry transformations to facilitate the recovery of inner cell structures $(\mathbf{A}_u, \mathbf{P}_u)$. We then represent atom $i$ in the \textbf{irreducible atom sets} by the atom type and fractional coordinates $[Z_i, \boldsymbol{p}_{\text{frac}, i}^x, \boldsymbol{p}_{\text{frac}, i}^y, \boldsymbol{p}_{\text{frac}, i}^z]$. It is worth noting that fractional coordinates of atom bases are invariant instead of equivariant, due to that $\mathbf{P} = \mathbf{L} \cdot \mathbf{P}_{\text{frac}}$ where $\cdot$ denotes matrix product and $\mathbf{R}\mathbf{P} = \mathbf{R} \mathbf{L} \cdot \mathbf{P}_{\text{frac}}$. The unique ordering of atoms within the irreducible set is determined by their atomic numbers, the number of duplicates recoverable from each atom, and their fractional coordinates, thereby ensuring consistency in the Mat2Seq sequence representations.

\subsection{Properties and proofs}
\label{sec:proofs}

\textbf{Uniqueness}. Different from previous crystal sequence representations~\cite{flam2023language, gruver2024fine, antunes2023crystal}, Mat2Seq sequence representations guarantee that all possible mathematical descriptions $\mathbf{M} = (\mathbf{A}, \mathbf{P}, \mathbf{L})$ for a given crystal will have a unique sequence representation and the same probability in the follow-up generative modeling process. We prove the uniqueness of Mat2Seq sequences as follows.

\begin{lemma}
A sequence mapping function $f:(\mathbf{A},\mathbf{P}, \mathbf{L}) \to \mathcal{X}$ is unique, if function $f$ is periodic invariant and unit cell $SE(3)$ invariant.
\end{lemma}
\begin{proof}
For a given crystal structure, a variety of structural descriptions $(\mathbf{A},\mathbf{P}, \mathbf{L})$ can be obtained by applying $SO(3)$ rotations $(\mathbf{A}, \bold{R}\mathbf{P}+b, \bold{R}\mathbf{L})$ and periodic transformations $(\mathbf{A}', \mathbf{P}', \mathbf{L}')$~\cite{comformer,cdvae} including (1) shifting periodic boundaries, and (2) altering periodic patterns while maintaining the same unit cell volume. 

For a sequence mapping function $f: (\mathbf{A},\mathbf{P}, \mathbf{L}) \to \mathcal{X}$ that satisfies periodic invariance and unit cell $SE(3)$ invariance, we have $f(\mathbf{A}, \mathbf{P}, \mathbf{L}) = f(\mathbf{A}, \bold{R}\mathbf{P}+b, \bold{R}\mathbf{L}) = f(\mathbf{A}', \mathbf{P}', \mathbf{L}')$ according to the definition provided in Sec.~\ref{sec:req}. Hence, a periodic invariant and unit cell $SE(3)$ invariant function is naturally unique for crystal structures. 
\end{proof}


\begin{lemma}
{Mat2Seq} $:(\mathbf{A},\mathbf{P}, \mathbf{L}) \to \mathcal{X}$ is periodic invariant and unit cell $SE(3)$ invariant.
\end{lemma}

\begin{proof}
    To prove that Mat2Seq is periodic invariant and unit cell $SE(3)$ invariant, we first show that the unit cell $(\mathbf{A}_u, \mathbf{P}_u, \mathbf{L}_u)$ determined in Sec.~\ref{sec:SO3_cell} is $SO(3)$ equivariant and periodic invariant. In other words, after applying arbitrary $SO(3)$ transformations $\bold{R} \in \mathbb{R}^{3 \times 3}, |\bold{R}|= 1$ with $\bold{b} \in \mathbb{R}^{3}$ and periodic transformations, the resultant unit cell should change from $(\mathbf{A}_u, \mathbf{P}_u, \mathbf{L}_u)$ to $(\mathbf{A}_u, \bold{R}\mathbf{P}_u, \bold{R}\mathbf{L}_u)$, with detailed proof provided in Appendix~\ref{app:proofs}.
     We then prove that the obtained sequence representation is $SE(3)$ and periodic invariant in Appendix~\ref{app:proofs}.  
\end{proof}

\textbf{Completeness}. Crystal sequence representations need to be complete to guarantee the full reconstruction of crystal structures from sequences.

\begin{lemma}
Crystal structure $(\mathbf{A},\mathbf{P}, \mathbf{L})$ can be fully recovered from {Mat2Seq} sequence $ \mathcal{X}$.
\end{lemma}

\begin{proof}
    $\mathbf{A}$ can be directly recovered from atom types in Mat2Seq sequences, and $\mathbf{L}$ can be recovered by three lattice lengths $a,b,c$ and three lattice angles $\alpha,\beta,\gamma$. Then, fractional coordinates of all atoms in the unit cell can be fully recovered by applying recorded space group transformations $[(\bold{R}_1, \bold{b}_1), \cdots, (\bold{R}_s, \bold{b}_s)]$ to fractional coordinates of the irreducible atom set $\mathbf{P}_{\text{frac}, \text{irr}}$ as follows,
    \begin{equation}
        \mathbf{P}_{\text{frac}} = (\mathbf{P}_{\text{frac}, \text{irr}} \oplus_{1\le i \le s} (\bold{R}_i \mathbf{P}_{\text{frac}, \text{irr}} + \bold{b}_i)~ \text{mod}~1)_{\text{set}},
    \end{equation}
    where $\oplus$ represents concatenation and $()_{\text{set}}$ represents the operation of removing duplicate entries.
    $\mathbf{P} = \mathbf{L} \cdot \mathbf{P}_{\text{frac}}$ can then be fully recovered.
\end{proof}

\begin{table}[t]
    \small
    \centering
    \caption{Uniqueness verification compared with previous crystal language models~\cite{flam2023language, gruver2024fine,antunes2023crystal}. }\vspace{-0.2cm}
    \begin{tabular}{c|ccc}
    \toprule
    Uniqueness - MP20~\cite{cdvae} & CIF~\cite{flam2023language, gruver2024fine} & CIF with symmetry~\cite{antunes2023crystal} & Mat2Seq \\
    \midrule
    Success rate $\uparrow$ & 0$\%$ & 30$\%$ & \textbf{100$\%$}\\
    \bottomrule
    \end{tabular}
    \label{tab:uniqueness}
\end{table}

\textbf{Remarks}. Based on \textbf{Lemma 1}, \textbf{2}, and \textbf{3}, Mat2Seq sequences satisfy uniqueness for crystal structures and can fully reconstruct corresponding 3D crystal structures. We also provide experimental verification on the MP20~\cite{cdvae} training set for the uniqueness of Mat2Seq and previous crystal language models~\cite{flam2023language, gruver2024fine,antunes2023crystal} in Table~\ref{tab:uniqueness} by applying periodic transformation that shifts the periodic boundaries. The evaluation is conducted as follows: (1) for each crystal structure in the MP20 dataset, we apply a transformation that shifts the periodic boundaries, generating a different unit cell representation while keeping the crystal structure unchanged; (2) we then encode both the original and transformed structures into 1D sequence representations using different methods and compare the resulting sequences. If there is a mismatch (e.g., differences in the coordinates or atom type for the first atom), it is considered a failure; (3) finally, we compute the success rate across the entire dataset.

\subsection{Language model enabled 3D crystal structure generation}
\label{sec:combine_LMs}

\textbf{Discretization}. To train a language model, crystal sequence representations obtained in Sec.~\ref{sec:SE3_seq} need to be discretized. Compositional and space group information including atom types, number of atoms in the cell, and space group symbol are discrete. For lattice parameters and fractional coordinates, we round real numbers to four decimal places to ensure consistency and efficiency following CrystaLLM~\cite{antunes2023crystal}. We also include keywords and symbols like space\_group\_symbol, formula, lattice\_parameters and so on in our Mat2Seq dictionary, with the full list provided in Appendix~\ref{app:csp}. 

\begin{table}[t]
\small
\centering
\caption{Comparison of Match Rate (\%) and RMSE across different models and crystal datasets.}\vspace{-0.2cm}
\label{tab:comparison}
\begin{tabular}{l|cc|cc|cc|cc}
\toprule
\multicolumn{1}{c|}{Model} & \multicolumn{2}{c|}{Perov-5} & \multicolumn{2}{c|}{Carbon-24} & \multicolumn{2}{c|}{MP20} & \multicolumn{2}{c}{MPTS-52} \\ 
& Match & RMSE & Match & RMSE & Match & RMSE & Match & RMSE \\ 
\midrule
& \multicolumn{8}{c}{One Shot} \\
\midrule
CrystaLLM & 46.1\% & \underline{0.095} & \underline{20.3\%} & \underline{0.176} & \underline{58.7\%} & \underline{0.041} & \underline{19.2\%} & \underline{0.111} \\
CDVAE & 45.3\% & 0.114 & 17.1\% & 0.297 & 33.9\% & 0.105 & 5.34\% & 0.211 \\
DiffCSP & \textbf{52.0\%} & \textbf{0.076} & 17.5\% & 0.276 & 51.5\% & 0.063 & 12.2\% & 0.179 \\
Mat2Seq & \underline{50.0\%} & 0.099 & \textbf{23.7\%} & \textbf{0.169} & \textbf{61.3\%} & \textbf{0.040} & \textbf{23.1\%} & \textbf{0.109} \\
\midrule
& \multicolumn{8}{c}{20 Shots} \\
\midrule
CrystaLLM & 97.6\% & 0.025 & 85.2\% & \underline{0.151} & 74.0\% & \textbf{0.035} & 33.8\% & \underline{0.106} \\
CDVAE & 88.5\% & 0.046 & \underline{88.4\%} & 0.229 & 67.0\% & 0.103 & 20.8\% & 0.209 \\
DiffCSP & \textbf{98.6\%} & \textbf{0.013} & \textbf{88.5\%} & 0.219 & \textbf{77.9\%} & 0.049 & \underline{34.0\%} & 0.175 \\
Mat2Seq & \underline{98.5\%} & \underline{0.023} & 86.0\% & \textbf{0.148} & \underline{75.3\%} & \underline{0.037} & \textbf{36.4\%} & \textbf{0.088} \\
\bottomrule
\end{tabular}
\end{table}

\textbf{Training and sampling}. We enable language models to generate novel crystals using the proposed unique and complete sequence representations for 3D crystal structures. Specifically, we follow the previous work~\cite{antunes2023crystal} and train a GPT~\cite{gpt} model $M$ with parameters $\theta$ to capture the distribution of Mat2Seq crystal sequences, $U$, constructed from corresponding crystal datasets with a standard next-token prediction cross-entropy loss $\ell$ for all elements in the sequence:
\begin{equation}
    \min_\theta \mathbb{E}_{u\in U} \sum_{i=1}^{|u|-1} \ell (M_\theta(u_1, u_2, \cdots, u_i), u_{i+1}),
\end{equation}
where $u$ represents a crystal structure within the dataset. During the sampling phase, we initiate the sequence with predetermined starting patterns, applying an autoregressive sampling technique to iteratively predict the next sequence component from the conditional model distribution $p_\theta(u_{i+1}|u_1,u_2,\cdots,u_i)$ until a stop pattern or a maximum length is achieved. Consistent with the referenced work~\cite{antunes2023crystal}, our sampling strategy incorporates top-$k$~\cite{topk} and temperature control~\cite{temp_control} techniques. Importantly, the choice of language model $M$ is flexible and can be seamlessly substituted with more advanced or alternative powerful language models.

\textbf{Conditional generation.} For conditional generation, we use conditional tokens to represent desired compositions and properties. Specifically, for Mat2Seq crystal sequence representations, the compositional information is placed ahead of crystal structure information. Hence, by simply initialize sequences with desired compositions, the language model $M$ can generate corresponding crystal structures. To target specific properties, we introduce a property token to denote the desired attribute and place these tokens at the beginning of the Mat2Seq sequences. The training and sampling processes mirror those used in unconditional generation, with the initial input being the property token that corresponds to the desired attribute.

\section{Experimental results}
\label{sec:exp_all}

In this section, we evaluate the ability of Mat2Seq to discover stable crystal structure for interested crystal systems in Sec.~\ref{sec:exp_1} and \ref{sec:discover_literature} and generate novel crystal structures satisfying desired physical properties in Sec.~\ref{sec:exp_prop}. We also include experimental results for random crystal generation assessing validity, stability (measured by DFT), and the ratios of stability, unique, and novel (S.U.N.) crystals in Appendix~\ref{app:add_exp}.

\subsection{Generating stable crystal structures for given compositions}
\label{sec:exp_1}

\textbf{Baselines}. We first evaluate the effectiveness of Mat2Seq on the widely used crystal structure prediction benchmark, comparing against diffusion-based methods including CDVAE~\cite{cdvae} and DiffCSP~\cite{jiao2024crystal}, as well as a recent language model based CrystaLLM~\cite{antunes2023crystal}.

\textbf{Datasets}. We conduct experiments on four datasets
following DiffCSP~\cite{jiao2024crystal} and CrystaLLM~\cite{antunes2023crystal}, including Perov-5, Carbon-24, MP-20, and MPTS-52. \textbf{Perov-5}~\cite{csp1,csp2} contains 18,928 perovskite materials with similar structures and 5 atoms in unit cells. \textbf{Carbon-24}~\cite{carbon} contains 10,153 carbon materials with at most 24 atoms in the unit cell. \textbf{MP-20}~\cite{jain2013commentary_mp} includes 45,231 stable inorgainic materials from the Materials Project, covering the majority of experimentally-generated materials with at most 20 atoms in the unit cell. \textbf{MPTS-52}~\cite{jiao2024crystal} is the most challenging one with 40,476 crystals and at most 52 atoms in the unit cell.

\begin{table}[t]
    \small
    \centering
    \caption{Efficiency and model complexity comparisons.}\vspace{-0.2cm}
    \begin{tabular}{c|cccc}
    \toprule
    MP-20 test set & number of parameters & RMSE & 20 shots generation speed (sec./crystal)\\
    \midrule
    CDVAE  & 4.5 M & 0.103  & 37.9 s \\
    DiffCSP  & Similar to CDVAE & 0.049  & 7.3 s \\
    Mat2Seq-small  & 25 M & \underline{0.039}  & \textbf{2.1 s} \\
    Mat2Seq-large  & 200 M & \textbf{0.037}  & \underline{5.7 s} \\
    \bottomrule
    \end{tabular}
    \label{tab:reb4_efficiency}
\end{table}

\begin{table}[t]
    \small
    \centering
    \caption{Mat2Seq match rate (\%) and RMSE for experimentally observed crystal structures in MP-20 test set.}\vspace{-0.2cm}
    \begin{tabular}{c|ccc}
    \toprule
    MP-20 & number of crystals & Match Rate (\%) & RMSE\\
    \midrule
    Mat2Seq - exp observed  & 3,819 & \textbf{65.2} \% & 0.042 \\
     Mat2Seq - the whole test set  & 9,046 & 61.3 \% & \textbf{0.040} \\
    \bottomrule
    \end{tabular}
    \label{tab:reb3}
    \vspace{-3mm}
\end{table}

\textbf{Experimental setup}. For crystal structure prediction, machine learning methods are trained using crystal structures and corresponding compositions in the training set. During the inference phase, compositions of crystals in the test set are provided as input to methods, and the goal is to generate stable crystal structures that match the ground truth crystal structures.  To assess the quality of generated crystal structures, two metrics are used, including match rate which measures the ratio of the generated structures that match with the ground truth structure determined by Pymatgen structure matcher~\cite{pymatgen}, and RMSE~\cite{pymatgen} which measures the structural differences between the ground truth and matched generated structures.
A single NVIDIA A100 GPU is used for computing for this task. We directly follow DiffCSP~\cite{jiao2024crystal} to split corresponding datasets into training, evaluation, and test sets. We use the same language model settings following CrystaLLM~\cite{antunes2023crystal} to make sure the comparison is fair. Detailed training and inference parameters are provided in Appendix~\ref{app:csp}. We also provide additional experimental evaluations in Appendix~\ref{app:add_exp} for random crystal generation on the MP20 dataset, assessing validity, stability (measured by DFT), and the ratios of stable, unique, and novel crystals, compared with state-of-the-art methods, including FlowMM~\cite{millerflowmm}, DiffCSP~\cite{jiao2024crystal}, and CDVAE~\cite{cdvae}.

\textbf{Results}. There are several observations from the experimental results shown in Table.~\ref{tab:comparison}. \textbf{(1)} Compared with CrystaLLM~\cite{antunes2023crystal} that directly uses CIF files to describe 3D crystal structures, Mat2Seq achieves better match rate for all eight tasks, demonstrating the effectiveness of Mat2Seq crystal sequence representations beyond CIF files. \textbf{(2)} Compared with DiffCSP~\cite{jiao2024crystal} that is specifically designed for this task, Mat2Seq achieves comparable results, and achieves a significantly better match rate and smaller RMSE for the most challenging MPTS-52 dataset, with a \textbf{89\%} improvement in match rate for one shot generation, and a \textbf{50\%} decrease in RMSE for 20 shots generation. \textbf{(3)} Mat2Seq demonstrates excellent one shot generation performances across datasets with various difficulty levels and data scales.

\textbf{Results on experimentally observed crystal structures}. Comparing with experimentally observed crystal structures is essential to demonstrate the real-world applicability of the proposed method, beyond synthetic structures generated purely from DFT calculations. The MP20 dataset test set includes 3,819 crystal structures that have been experimentally observed (match entries in the ICSD~\cite{zagorac2019recent}). To assess the accuracy of Mat2Seq on this data, we conducted additional evaluations on these 3,819 structures, with the results presented in Table~\ref{tab:reb3}. Notably, Mat2Seq achieved a 65.2\% match rate with an RMSE of 0.042 for these experimentally observed structures.

\textbf{Efficiency}. It is well known that language models typically have higher parameter counts and greater computational demands. To address this, we provide a comparison of efficiency in terms of parameter count and generation speed, alongside specialized crystal models such as CDVAE and DiffCSP, in Table~\ref{tab:reb4_efficiency}. While Mat2Seq, based on large language models, does have more parameters, using similar GPU resources (as done in DiffCSP~\cite{jiao2024crystal}), Mat2Seq demonstrates significantly faster generation speed. Furthermore, we show that reducing the model size by a factor of eight does not significantly affect Mat2Seq's performance and improves efficiency. In this case, the smaller model achieved an RMSE of 0.039, which is notably better than DiffCSP's RMSE of 0.049, while also being more than three times faster in the generation process.

\subsection{Ability to generalize to recently discovered crystals from literature}
\label{sec:discover_literature}

The promising results in crystal structure prediction shown in Sec.~\ref{sec:exp_1} have demonstrated the effectiveness of Mat2Seq crystal sequence representations. We then examine the ability of Mat2Seq with language model to discover novel crystal structures that are obtained from recent literature and are not seen during training. Specifically, we randomly choose 10 crystals from the \textbf{challenge set}~\cite{antunes2023crystal} to evaluate Mat2Seq's ability to discover novel crystal structures that originate from recent literature, and compare with previous CrystaLLM~\cite{antunes2023crystal}. 

\textbf{Dataset}. For this task, we follow CrystaLLM~\cite{antunes2023crystal} and train the model using crystal structures from the Materials Project~\cite{jain2013commentary_mp}, OQMD~\cite{saal2013materials} and NOMAD~\cite{scheidgen2023nomad}, which has been optimized using DFT calculations, containing 2,047,889 crystal structures for training in total. This dataset contains around 800K unique formulas and 94 chemical element types. The evaluation is conducted with 10 recently discovered crystal structures from literature, including AlCu$_2$As(OH)$_{12}$~\cite{AlCu2}, Ba$_2$Fe$_2$F$_9$~\cite{Ba2Fe2}, Ca$_2$Te$_3$O$_8$~\cite{Ca2Te3}, $\text{Ce}_{6}\text{Cd}_{23}\text{Te}$~\cite{Ce6Cd23}, Cs$_8$Cu$_3$Si$_{14}$O${_{35}}$~\cite{Cs8Cu3}, Cu$_4$FeGe$_2$OS$_6$~\cite{Cu4Fe}, Eu$_2$FeGe$_2$OS$_6$~\cite{Eu2}, K$_2$Sr$_4$(PO$_3$)$_{10}$~\cite{K2Sr4}, La$_4$Ga$_2$S$_8$O$_3$~\cite{La4Ga2}, and Li$_9$Al$_4$Sn$_5$~\cite{Li9Al4}.

\textbf{Experimental setup}. We first train our model using around 2M crystal structures in the training set following CrystaLLM with broken files removed. After training, we evaluate the model's ability to generate valid crystal structures for 10 recently discovered crystal structures from literature by taking the compositional information as conditions and generating 20 structures for each composition. To compare with CrystaLLM, we directly use their pre-trained model for the same training set and generate 20 structures for each composition. We follow CrystaLLM and evaluate the validity of generated structures by whether the generated formula is consistent with the given composition, whether the space group is consistent with the structure, and whether the generated crystal has reasonable bond lengths. 4 A100 GPUs are used for computing for this task. We also provide additional experimental results in Appendix~\ref{app:add_exp}, evaluating the Hit rate (i.e., whether the generated structure matches recently observed experimental crystals from the literature) and RMSE across 10 challenge crystal systems, in comparison with the previous state-of-the-art method, CrystaLLM.

\begin{figure}[t]
    \centering\includegraphics[width=0.99\linewidth]{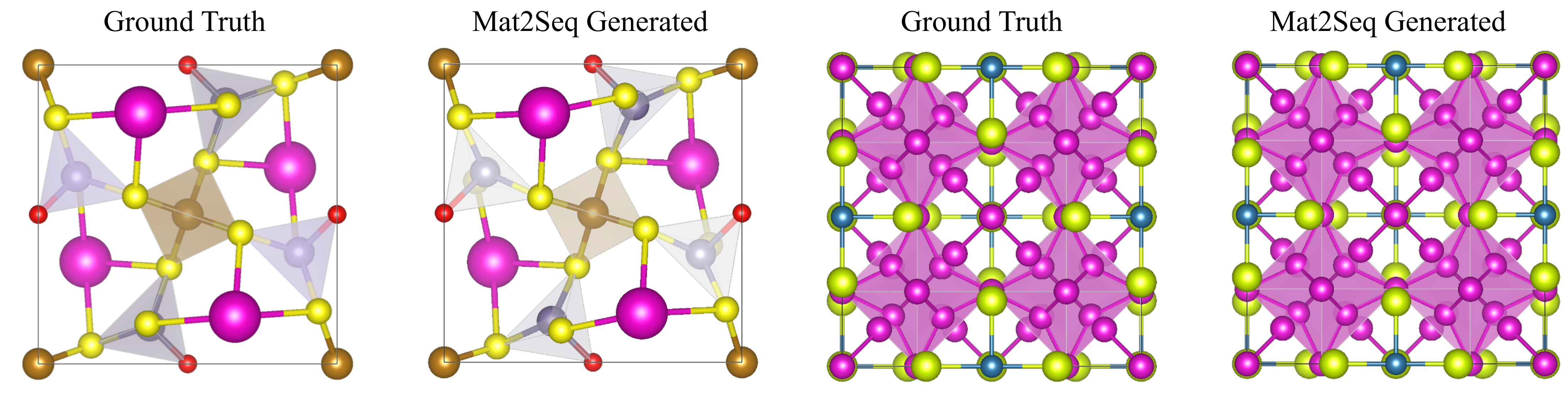}\vspace{-0.3cm}
    \caption{Mat2Seq can generate recently discovered novel crystals from literature. Eu$_2$FeGe$_2$OS$_6$ on the left, and $\text{Ce}_{6}\text{Cd}_{23}\text{Te}$ on the right. The structure generated by Mat2Seq for Eu$_2$FeGe$_2$OS$_6$ is the reflected version of the ground truth.}
    \label{fig:mat2seq_generated_cifs}
\end{figure}

\begin{table}[]
    \centering
    \caption{Ability to generalize to recently discovered crystals from literature, measured by three seperate generation runs.}
    \begin{tabular}{c|cc}
    \toprule
    &CrystaLLM& Ours \\
    \midrule
    Validity $\uparrow$ $\pm$ std & 23.7$\%$ $\pm$ 2.5$\%$ & \textbf{27.3$\%$}  $\pm$ 1.4$\%$  \\
    \bottomrule
    \end{tabular}
    \label{tab:validity}
\end{table}


\textbf{Results}. As shown in Table~\ref{tab:validity}, for the 20$\times$10 generated crystals, our method achieved \textbf{27.3\%} validity rate for novel crystal structures from recent literature, better than CrystaLLM with 23.7\%. We visualize the generation results for Eu$_2$FeGe$_2$OS$_6$ and $\text{Ce}_{6}\text{Cd}_{23}\text{Te}$ in Figure~\ref{fig:mat2seq_generated_cifs} which shows Mat2Seq actually can discover novel crystals reported in recent literature with nearly identical 3D structures. It is worth noting that the structure generated by Mat2Seq for Eu$_2$FeGe$_2$OS$_6$ is the reflected version of the founded one, and reflected crystal structure will have the identical energy, which on the other hand demonstrating the reflection sensitivity of Mat2Seq.

\subsection{Discovering crystal structures with desired properties}
\label{sec:exp_prop}

\textbf{Experimental details}. We further evaluate Mat2Seq's ability to discover crystals with desired properties. Specifically, we have put 10 placeholders marked as unknown\_prop before Mat2Seq crystal sequences when training our model on around 2M crystal structures. After pre-training, we fine-tune the model on JARVIS-DFT dataset~\cite{jarvis} with 61,541 crystal structures and corresponding band gap values in the training set. We use intervals to tokenize band gap values, where values from 0 to 0.5 are marked as 0, values from 0.5 to 1 are marked as 1, and so on, and replace one unknown\_prop token in the sequence. For future users, this design can enable the discovery of physical properties of their interest by simply replacing the unknown\_prop token with corresponding properties and fine-tuning the pre-trained model from Sec.~\ref{sec:discover_literature}. A single A100 GPU is used for this task.

\textbf{Evaluation metrics}. We evaluate our model in two settings, including (1) generating crystals with low band gap ($< 0.5$ eV) that could potentially be used for thermoelectrics, thermophotovoltaics, infrared sensing and vision, cryogenic cooling, as well as discovering novel catalysts and topological quantum materials, and (2) generating crystals with high band gap ($> 3.0$ eV) that could potentially be used for solid-state LEDs and lasers, nonlinear optics, ferroelectrics, multiferroics, power electronics, and high-temperature applications.  We measure the success rate of 500 randomly generated crystals that satisfy (1) and (2) when our model is conditioned towards generating crystals with small band gap values and large band gap values. We measure the band gap value of generated crystals using the state-of-the-art method ComFormer~\cite{comformer} for band gap prediction with mean absolute error of 0.122 eV. We also provide measurements of validity, uniqueness, and novelty for the generated crystals.

\textbf{Results}. Table~\ref{tab:prop_cond} demonstrates Mat2Seq's ability to significantly alter the original property distribution and discover crystals towards large or small band gap values. Specifically, when conditioned towards small band gap values with interval 0 to 0.5 eV, \textbf{83.6\%} generated crystals have band gap values $<0.5$ eV, and when conditioned towards large band gap values with interval 3.5 to 4.0 eV, \textbf{90.7\%} generated crystals have band gap values $>3.0$ eV.

\begin{table}[t]
    \small
    \centering
    \caption{Ability to generate crystal structures with desired band gap properties. We measure the success rate of generating crystal structures with band gap value < 0.5 eV and band gap value > 3.0 eV, measured by the state-of-the-art band gap predictor ComFormer~\cite{comformer}. We also evaluate the validity, uniqueness, and novelty of the generated crystals.}\vspace{-0.2cm}
    \begin{tabular}{c|ccccc}
    \toprule
    Generation Codition & band gap < 0.5 eV & band gap > 3.0 eV & Validity & Uniqueness & Novelty \\
    \midrule
     Towards band gap $\downarrow$ & \textbf{83.6$\%$} & 12.0$\%$ & 88.0$\%$ & \textbf{98.0}$\%$& 86.2$\%$\\
     Towards band gap $\uparrow$ & 6.4$\%$ & \textbf{90.7$\%$}& \textbf{89.8}$\%$ & 92.2$\%$& \textbf{98.6}$\%$ \\
    \bottomrule
    \end{tabular}
    \label{tab:prop_cond}
    \vspace{-3mm}
\end{table}

\section{Conclusion, limitation, and discussion}
\label{sec:conclusion}

In this work, we propose Mat2Seq, which converts 3D crystals into unique and complete 1D sequences for language model-enabled crystal generation, naturally ensuring that all equivalent crystals are considered identically probable without the need for data augmentation. Mat2Seq demonstrates promising performance in crystal structure generation and discovery, and the trained model can be used to discover crystal structures with desired properties of interest to users. The limitations of the current Mat2Seq include: (1) it cannot be directly used for other atomic systems, like molecules and proteins; (2) the extension to model disordered materials remains a challenging frontier; and (3) large-scale training with more stable crystal structures can potentially enhance the robustness and performance when more computational resources are available. Positive and negative societal impacts of discovering novel materials with desired properties may apply to this work.

\begin{ack}

K.Y. and S.J. acknowledge the support from U.S. National Science Foundation~(NSF) grant MOMS-2331036.
First-principles calculations and structure optimization by K.A. were supported by the Center for Reconfigurable Electronic Materials Inspired by Nonlinear Dynamics (reMIND), an Energy Frontier Research Center funded by the Department of Energy under award DE-SC0023353. X.F.Q. acknowledges the support from the Air Force Office of Scientific Research (AFOSR) under Grant No. FA9550-24-1-0207 and NSF CMMI-2226908. This work was partially supported by the donors of ACS Petroleum Research Fund under Grant 65502-ND10. X.N.Q. and R.A. acknowledge the support from U.S. National Science Foundation~(NSF) grant DMREF-2119103 and X.N.Q acknowledges the support from NSF through grants SHF-2215573, and IIS-2212419. Portions of this research were conducted with the advanced computing resources provided by Texas A\&M High Performance Research Computing.
M.Z.~gratefully acknowledges the support of NIH R01-HD108794, NSF CAREER 2339524, US DoD FA8702-15-D-0001, ARPA-H BDF Toolbox, awards from Pfizer Research, Harvard Data Science Initiative, Amazon Faculty Research, Google Research Scholar Program, AstraZeneca Research, Roche Alliance with Distinguished Scientists, Sanofi iDEA-iTECH Award, Chan Zuckerberg Initiative, John and Virginia Kaneb Fellowship award at Harvard Medical School, Biswas Computational Biology Initiative in partnership with the Milken Institute, Kempner Institute for the Study of Natural and Artificial Intelligence, and Harvard Medical School Dean’s Innovation Fund.
C.E. and H.J. acknowledge the support from the Molecule Maker Lab Institute: an AI research institute program supported by NSF under award No. 2019897 and No. 2034562. The views and conclusions contained herein are those of the authors and should not be interpreted as necessarily representing the official policies, either expressed or implied, of the U.S. Government. The U.S. Government is authorized to reproduce and distribute reprints for governmental purposes notwithstanding any copyright annotation therein.

\end{ack}

\bibliographystyle{unsrt}
\bibliography{dive}


\newpage

\appendix

\section{Appendix}

\subsection{Proofs}
\label{app:proofs}

\begin{lemma}
{Mat2Seq} $:(\mathbf{A},\mathbf{P}, \mathbf{L}) \to \mathcal{X}$ is periodic invariant and unit cell $SE(3)$ invariant.
\end{lemma}

\begin{proof}
    To prove that Mat2Seq is periodic invariant and unit cell $SE(3)$ invariant, we first show that the unit cell $(\mathbf{A}_u, \mathbf{P}_u, \mathbf{L}_u)$ determined in Sec.~\ref{sec:SO3_cell} is $SO(3)$ equivariant and periodic invariant. In other words, after applying arbitrary $SO(3)$ transformations $\bold{R} \in \mathbb{R}^{3 \times 3}, |\bold{R}|= 1$ with $\bold{b} \in \mathbb{R}^{3}$ and periodic transformations, the resultant unit cell should change from $(\mathbf{A}_u, \mathbf{P}_u, \mathbf{L}_u)$ to $(\mathbf{A}_u, \bold{R}\mathbf{P}_u, \bold{R}\mathbf{L}_u)$.
    To begin with, because Euclidean distances $a,b,c$, bond angles $\alpha,\beta,\gamma$, and whether three vectors form a right-hand system are $SE(3)$ and periodic invariant, which means after arbitrary $SO(3)$ transformation $\bold{R} \in \mathbb{R}^{3 \times 3}, |\bold{R}|= 1$ with $\bold{b} \in \mathbb{R}^{3}$ and periodic transformation, the same but rotated lattice matrix $\mathbf{R}\mathbf{L}_u$ will be determined. 
    Additionally, because local density $\rho_{i,r}$, atom type $Z_i$, and densities along three lattice vectors $\rho_{i,r}^{\boldsymbol{\ell}_1'}, \rho_{i,r}^{\boldsymbol{\ell}_2'}, \rho_{i,r}^{\boldsymbol{\ell}_3'}$ are $SE(3)$ and periodic invariant,
    the same atom $i$ will be determined to serve as the origin of the unit cell after arbitrary $SO(3)$ transformation and periodic transformation. Hence, when applying arbitrary $SO(3)$ transformation $\bold{R} \in \mathbb{R}^{3 \times 3}, |\bold{R}|= 1$, $\bold{b} \in \mathbb{R}^{3}$ and periodic transformation to $(\mathbf{A}, \mathbf{P}, \mathbf{L})$, the resultant 
    primitive unit cell will be $(\mathbf{A}_u, \bold{R}\mathbf{P}_u, \bold{R}\mathbf{L}_u)$. Hence, the unit cell $(\mathbf{A}_u, \mathbf{P}_u, \mathbf{L}_u)$ determined by Mat2Seq is $SO(3)$ equivariant and periodic invariant.
    
    Based on the $SO(3)$ equivariant and periodic invariant primitive unit cell $(\mathbf{A}_u, \mathbf{P}_u, \mathbf{L}_u)$, we then prove that the obtained sequence representation is $SE(3)$ and periodic invariant. To begin with, it can be seen that lattice lengths, lattice angles, and atomic numbers are naturally invariant. Additionally, due to that $\mathbf{P}_u, \mathbf{L}_u$ in  $(\mathbf{A}_u, \mathbf{P}_u, \mathbf{L}_u)$ are $SO(3)$ equivariant, periodic invariant, and are connected by fractional coordinates $\boldsymbol{p}_i = \boldsymbol{p}_{ \text{frac},i} \cdot \mathbf{L} = x *\boldsymbol{\ell}_1 + y * \boldsymbol{\ell}_2 + z *\boldsymbol{\ell}_3 $, when the crystal is rotated by $\mathbf{R}$ we can have 
    \begin{align*}
        \mathbf{R}\boldsymbol{p}_i = \mathbf{R} (x *\boldsymbol{\ell}_1 + y * \boldsymbol{\ell}_2 + z *\boldsymbol{\ell}_3) = x' * \mathbf{R}\boldsymbol{\ell}_1 + y' * \mathbf{R}\boldsymbol{\ell}_2 + z' *\mathbf{R}\boldsymbol{\ell}_3,
    \end{align*}
    which means $x = x', y = y', z=z'$. Hence fractional coordinates are $SE(3)$ invariant. Additionally, because $\mathbf{P}_u, \mathbf{L}_u$ are periodic invariant, the corresponding fractional coordinates are periodic invariant. To summarize, it can be seen that all components in Mat2Seq sequence, including (1) compositional information, (2) space group information, (3) lattice parameters obtained by lattice lengths and angles, and (4) atom types and fractional coordinates are $SE(3)$ and periodic invariant. 
\end{proof}

\subsection{Experimental details}
\label{app:csp}

Following CrystaLLM~\cite{antunes2023crystal}, we use GPT-2 with 16 layers, 16 heads, and embedding size of 1024 for all tasks. We show the detailed training parameters including window size, batch size, learning rate, drop out ratio, number of training iterations for different tasks in Table.~\ref{tab:app_para}. 

\begin{table}[h]
\centering
\caption{Training parameters of Mat2Seq for crystal structure prediction benchmark.}
\label{tab:app_para}
\begin{tabular}{c|ccccc}
\toprule
Task &  Window size & Batch size & Learning rate & Num. iterations & Drop out \\
\midrule
Perov-5 & 768 & 32 & 0.0005 & 40k & 0.1 \\
Carbon-24 & 768 & 48 & 0.0003 & 40k & 0.1\\
MP20 & 768 & 32 & 0.0003 & 50k & 0.1 \\
MPTS52 & 1,385 & 24 & 0.0003 & 40k & 0.1 \\
CrystaLLM 2.3 M & 1,500 & 128 & 0.0005 & 500k & 0.1 \\
JARVIS bandgap & 1,500 & 20 & 0.0001 & 10k & 0.1 \\
\bottomrule
\end{tabular}
\end{table}

During sampling phase, for Perov-5 dataset, we use temperature=0.7 and top-$k$=10 one shot generation, and temperature=1.0, top-$k$=10 for 20 shots generation. For Carbon-24 dataset, we use temperature=0.7 and top-$k$=10 one shot generation, and temperature=1.0, top-$k$=10 for 20 shots generation. For MP20 dataset, we use temperature=0.4 and top-$k$=5 one shot generation, and temperature=1.5, top-$k$=10 for 20 shots generation. 
for MPTS52 dataset, we use temperature=1.0 and top-$k$=10 one shot generation, and temperature=1.0, top-$k$=10 for 20 shots generation.
For discovering novel crystal structures obtained from recent literature, we use temperature=0.7 and top-$k$=10 during sampling. For discovering crystal structures with desired band gap properties, we use temperature=0.8 and top-$k$=10 during sampling phases.

\textbf{Mat2Seq tokens}. We provide comprehensive list of tokens used by Mat2Seq, including atom types listed in Table~\ref{tab:atom_type_list}, integer numbers from 0 to 300, digits including 0 to 9 with "." to represent real values, special tokens listed in Table~\ref{tab:special_token}, and 227 space group symbols~\cite{antunes2023crystal}.

\begin{table}[h!]
  \centering
  \caption{List of atom types.}
  \begin{tabular}{cccccccccccc}
    \toprule
    Ac & Ag & Al & Ar & As & Au & B & Ba & Be & Bi & Br & C \\
    Ca & Cd & Ce & Cl & Co & Cr & Cs & Cu & Dy & Er & Eu & F \\
    Fe & Ga & Gd & Ge & H & He & Hf & Hg & Ho & I & In & Ir \\
    K & Kr & La & Li & Lu & Mg & Mn & Mo & N & Na & Nb & Nd \\
    Ne & Ni & Np & O & Os & P & Pa & Pb & Pd & Pm & Pr & Pt \\
    Pu & Rb & Re & Rh & Ru & S & Sb & Sc & Se & Si & Sm & Sn \\
    Sr & Ta & Tb & Tc & Te & Th & Ti & Tl & Tm & U & V & W \\
    Xe & Y & Yb & Zn & Zr \\
    \bottomrule
  \end{tabular}
  \label{tab:atom_type_list}
\end{table}

\begin{table}[h!]
  \centering
  \caption{List of special tokens.}
  \begin{tabular}{ccccccc}
    \toprule
    space\_group\_symbol & formula & atoms & lattice\_parameters & a & b & c \\ alpha & beta & gamma & unknown\_prop & , & " " & :\\ \textbackslash n & <pad>\\
    \bottomrule
  \end{tabular}
  \label{tab:special_token}
\end{table}

\subsection{Additional Experimental Results}
\label{app:add_exp}

We provide additional experimental results of Mat2Seq in this section. 

In Table~\ref{tab:reb_1}, we compare Mat2Seq with the recent state-of-the-art method FlowMM~\cite{millerflowmm}, assessing Validity, Stability, and S.U.N. (stable, unique, and novel) metrics. To ensure a fair comparison, we follow the FlowMM pipeline and generate 1000 materials to obtain these results.

\begin{table}[h]
\small
\centering
\caption{Comparison of Validity (\%), Stability (\%). and Stable, Unique, and Novel (S.U.N) (\%) on MP20 dataset. The unit of $E_{hull
}$ is eV/atom.}\vspace{-0.2cm}
\label{tab:reb_1}
\begin{tabular}{lccccc}
\toprule
\multicolumn{1}{c}{Model} & \multicolumn{2}{c}{Validity ($\%$) $\uparrow$} & \multicolumn{2}{c}{Stability Rate - DFT ($\%$) $\uparrow$} & \multicolumn{1}{c}{S.U.N. Rate ($\%$) $\uparrow$} \\ 
& Composition & Structural & $E_{hull
} < 0.0$ & $E_{hull
} < 0.1$ & MP  \\ 
\midrule
CDVAE & \underline{86.7} \% & \textbf{100} \% & 1.57\% & - & 1.4\%\\
DiffCSP & 83.3 \% & \textbf{100} \% & \textbf{5.06}\% & - & \textbf{3.3}\% \\
FlowMM (ICML 24)& 83.2 \% & 96.9 \% & 4.19\% & - & 2.5\%\\
\midrule
Mat2Seq (temp=1.35) & \textbf{88.5} \% & 94.2 \% & 4.10\% & \textbf{49.2}\% & 2.0\%\\
Mat2Seq (temp=1.65) & 81.7 \% & 88.6 \% & \underline{4.50}\% & \underline{46.6}\% & \underline{3.2}\% \\
\bottomrule
\end{tabular}
\end{table}


In Table~\ref{tab:reb_2}, we present the Hit Rate (whether the generated structures match the ground truth structure) and RMSE for the 10 challenge crystal systems discussed in Section~\ref{sec:discover_literature}, compared to the previous state-of-the-art method, CrystaLLM.

\begin{table}[h!]
\small
\centering
\caption{Comparison of Hit Rate ($\%$) and RMSE for ten recently discovered crystal structures from literature.}\vspace{-0.2cm}
\label{tab:reb_2}
\begin{tabular}{lcc}
\toprule
Model & Hit Rate ($\%$) $\uparrow$ & RMSE $\downarrow$ \\ 
\midrule
CrystaLLM & \textcolor{red}{0} & \textcolor{red}{nan}\\
Mat2Seq & \textbf{10.0 $\%$} & \textbf{0.0388}\\
\bottomrule
\end{tabular}
\end{table}

\subsection{Licenses for existing assets}
\label{app:license}

We have used datasets including Perov-5, Carbon-24, and MP20 curated by CDVAE~\cite{cdvae} with MIT License, MPTS-52 curated by DiffCSP~\cite{jiao2024crystal} with MIT License, JARVIS-DFT~\cite{jarvis} with NIST License, CrystaLLM~\cite{antunes2023crystal} with MIT License,  the Materials Project~\cite{jain2013commentary_mp} with Creative Commons Attribution 4.0 License, OQMD~\cite{saal2013materials} with Creative Commons Attribution 4.0 International License, and NOMAD~\cite{scheidgen2023nomad} with Apache License Version 2.0, January 2004. We have used CrystaLLM~\cite{antunes2023crystal} model with MIT License.


\newpage
\section*{NeurIPS Paper Checklist}

\begin{enumerate}

\item {\bf Claims}
    \item[] Question: Do the main claims made in the abstract and introduction accurately reflect the paper's contributions and scope?
    \item[] Answer: \answerYes{} 
\item[] Justification: As clearly discussed in Sec.~\ref{sec:method} and Sec.~\ref{sec:exp_all}, the main claims made in the abstract and introduction accurately reflect the paper's contributions and scope.

\item {\bf Limitations}
    \item[] Question: Does the paper discuss the limitations of the work performed by the authors?
    \item[] Answer: \answerYes{} 
    \item[] Justification: Limitations of this work are discussed in Sec.~\ref{sec:conclusion}.

\item {\bf Theory Assumptions and Proofs}
    \item[] Question: For each theoretical result, does the paper provide the full set of assumptions and a complete (and correct) proof?
    \item[] Answer: \answerYes{} 
    \item[] Justification: Detailed proofs are provided in Sec.~\ref{sec:proofs} and Appendix~\ref{app:proofs}.

    \item {\bf Experimental Result Reproducibility}
    \item[] Question: Does the paper fully disclose all the information needed to reproduce the main experimental results of the paper to the extent that it affects the main claims and/or conclusions of the paper (regardless of whether the code and data are provided or not)?
    \item[] Answer: \answerYes{} 
    \item[] Justification: Detailed experimental settings, datasets, and computations needed are shared in Sec.~\ref{sec:exp_all} and Appendix~\ref{app:csp}.

\item {\bf Open access to data and code}
    \item[] Question: Does the paper provide open access to the data and code, with sufficient instructions to faithfully reproduce the main experimental results, as described in supplemental material?
    \item[] Answer: \answerYes{} 
    \item[] Justification: The code will be release after the paper is publicly available.

\item {\bf Experimental Setting/Details}
    \item[] Question: Does the paper specify all the training and test details (e.g., data splits, hyperparameters, how they were chosen, type of optimizer, etc.) necessary to understand the results?
    \item[] Answer: \answerYes{} 
    \item[] Justification: Detailed experimental settings are provided in Sec.~\ref{sec:exp_all} and Appendix~\ref{app:csp}.

\item {\bf Experiment Statistical Significance}
    \item[] Question: Does the paper report error bars suitably and correctly defined or other appropriate information about the statistical significance of the experiments?
    \item[] Answer: \answerYes{} 
    \item[] Justification: We provide the validity of randomly generated crystal structures for interested crystal compositions from recent literature, including 20 generation runs for each composition. We also repeat this generation process for our method and the baseline method three times and show the std in Sec.~\ref{sec:exp_all}. For crystal structure prediction tasks, we follow previous benchmark settings and include 20 shots generation results. 

\item {\bf Experiments Compute Resources}
    \item[] Question: For each experiment, does the paper provide sufficient information on the computer resources (type of compute workers, memory, time of execution) needed to reproduce the experiments?
    \item[] Answer: \answerYes{} 
    \item[] Justification: Detailed compute resources needed are provided in Sec.~\ref{sec:exp_all}.
    
\item {\bf Code Of Ethics}
    \item[] Question: Does the research conducted in the paper conform, in every respect, with the NeurIPS Code of Ethics \url{https://neurips.cc/public/EthicsGuidelines}?
    \item[] Answer: \answerYes{} 
    \item[] Justification: We conform that we follow NeurIPS Code of Ethics in every respect.

\item {\bf Broader Impacts}
    \item[] Question: Does the paper discuss both potential positive societal impacts and negative societal impacts of the work performed?
    \item[] Answer: \answerYes{} 
    \item[] Justification: We provide Broader Impacts in Sec.~\ref{sec:conclusion}.
    
\item {\bf Safeguards}
    \item[] Question: Does the paper describe safeguards that have been put in place for responsible release of data or models that have a high risk for misuse (e.g., pretrained language models, image generators, or scraped datasets)?
    \item[] Answer: \answerNo{} 
    \item[] Justification: We will consider this when we release the model in the future.

\item {\bf Licenses for existing assets}
    \item[] Question: Are the creators or original owners of assets (e.g., code, data, models), used in the paper, properly credited and are the license and terms of use explicitly mentioned and properly respected?
    \item[] Answer: \answerYes{} 
    \item[] Justification: We cite the existing assets in the main paper and provide licenses of them in Appendix~\ref{app:license}.

\item {\bf New Assets}
    \item[] Question: Are new assets introduced in the paper well documented and is the documentation provided alongside the assets?
    \item[] Answer: \answerYes{} 
    \item[] Justification: We discuss details of the dataset and model in Sec.~\ref{sec:exp_all} and Appendix~\ref{app:csp}.

\item {\bf Crowdsourcing and Research with Human Subjects}
    \item[] Question: For crowdsourcing experiments and research with human subjects, does the paper include the full text of instructions given to participants and screenshots, if applicable, as well as details about compensation (if any)? 
    \item[] Answer: \answerNA{} 
    \item[] Justification: This paper does not involve crowdsourcing nor research with human subjects.

\item {\bf Institutional Review Board (IRB) Approvals or Equivalent for Research with Human Subjects}
    \item[] Question: Does the paper describe potential risks incurred by study participants, whether such risks were disclosed to the subjects, and whether Institutional Review Board (IRB) approvals (or an equivalent approval/review based on the requirements of your country or institution) were obtained?
    \item[] Answer: \answerNA{} 
    \item[] Justification: This paper does not involve crowdsourcing nor research with human subjects.

\end{enumerate}

\newpage


\end{document}